\pdfoutput=1

\documentclass[11pt]{article}

\usepackage{amsthm}
\usepackage{amssymb}
\usepackage{mathtools}
\usepackage{hyperref}
\usepackage[nameinlink,capitalize]{cleveref}
\hypersetup{colorlinks=true,linkcolor=blue,citecolor=blue,urlcolor=blue,pdfborder={0 0 0}}
\usepackage[normalem]{ulem} %
\usepackage{mathtools}
\usepackage{algorithm}
\usepackage{algorithmic}

\usepackage[utf8]{inputenc} %
\usepackage[T1]{fontenc}    %
\usepackage{hyperref}       %
\usepackage{url}            %
\usepackage{booktabs}       %
\usepackage{amsfonts}       %
\usepackage{nicefrac}       %
\usepackage{microtype}      %
\usepackage{color, colortbl}
\definecolor{Gray}{gray}{0.9}
\newcolumntype{g}{>{\columncolor{Gray}}c}
\usepackage{cancel}
\usepackage[normalem]{ulem}

\definecolor{Gray}{gray}{0.93}
\newcolumntype{a}{>{\columncolor{Gray}}c}

\usepackage{graphicx,wrapfig}
\usepackage{caption}
\usepackage{amsfonts}
\usepackage{url}
\usepackage{amsthm}
\usepackage{thmtools}
\usepackage{thm-restate}

\newcommand{\R}{\mathbb{R}}

\newcommand{\st}{{\text{s.t.}}} %
\newcommand{\diag}{{\mathrm{diag}}} %
\newcommand{\tr}{{\mathrm{tr}}} %
\DeclareMathOperator*{\argmin}{arg\,min}

\newcommand{\bc}{\begin{center}}
\newcommand{\ec}{\end{center}}

\newcommand{\bdm}{\begin{displaymath}}
\newcommand{\edm}{\end{displaymath}}

\newcommand{\beq}{\begin{equation}}
\newcommand{\eeq}{\end{equation}}

\newcommand{\bfl}{\begin{flushleft}}
\newcommand{\efl}{\end{flushleft}}

\newcommand{\bt}{\begin{tabbing}}
\newcommand{\et}{\end{tabbing}}

\newcommand{\beqn}{\begin{align}}
\newcommand{\eeqn}{\end{align}}

\newcommand{\beqs}{\begin{align*}} %
\newcommand{\eeqs}{\end{align*}}  %

\usepackage[final]{acl}

\usepackage{times}
\usepackage{latexsym}

\usepackage[T1]{fontenc}

\usepackage[utf8]{inputenc}

\usepackage{microtype}

\usepackage{inconsolata}

\usepackage{graphicx}
\usepackage{comment}
\usepackage{amsfonts}
\usepackage{amssymb}
\usepackage{amsmath}
\usepackage{multirow}
\usepackage{subfigure}

\title{Unraveling LoRA Interference: Orthogonal Subspaces for Robust Model Merging}

\author{Haobo Zhang \\
  University of Michigan \\
  Ann Arbor, USA \\
    \texttt{haobozha@umich.edu} 
    \\\And
  Jiayu Zhou \\
  University of Michigan \\
  Ann Arbor, USA \\
  \texttt{jiayuz@umich.edu} 
  \\}

\begin{document}
\maketitle

\begin{abstract}
    Fine-tuning large language models (LMs) for individual tasks yields strong performance but is expensive for deployment and storage. 
    Recent works explore model merging to combine multiple task-specific models into a single multi-task model without additional training. 
    However, existing merging methods often fail for models fine-tuned with low-rank adaptation (LoRA), due to significant performance degradation. 
    In this paper, we show that this issue arises from a previously overlooked interplay between model parameters and data distributions. 
    We propose \textbf{O}rthogonal \textbf{S}ubspaces for \textbf{R}obust model \textbf{M}erging (\textbf{OSRM}) to constrain the LoRA subspace \emph{prior} to fine-tuning, ensuring that updates relevant to one task do not adversely shift outputs for others. 
    Our approach can seamlessly integrate with most existing merging algorithms, reducing the unintended interference among tasks. 
    Extensive experiments on eight datasets, tested with three widely used LMs and two large LMs, demonstrate that our method not only boosts merging performance but also preserves single-task accuracy.
    Furthermore, our approach exhibits greater robustness to the hyperparameters of merging. 
    These results highlight the importance of data-parameter interaction in model merging and offer a plug-and-play solution for merging LoRA models.
\end{abstract}

\section{Introduction}
\label{sec:intro}

Pre-trained language models (LMs) have achieved remarkable success across diverse tasks, with fine-tuning approaches enabling strong downstream performance \citep{gpt2,llama2}.
However, maintaining a separate fine-tuned model for each task becomes prohibitively expensive in terms of both storage and deployment. 
While multi-task learning \citep{multitask} attempts to address this issue by training a unified model for multiple tasks, it demands simultaneous access to all task data and high computational overhead, thereby limiting its scalability.

An appealing alternative is model merging, which combines multiple task-specific models into a single multi-task model without further training \citep{taskarithmetic,emr,ties}. 
Early merging techniques primarily average parameters, often guided by Fisher information \citep{fisher} or inner-product-based metrics \citep{regmean}. 
Another line of work employs task vectors---the difference between pre-trained and fine-tuned model weights---and manipulates them 
before summation \citep{ties,pcb,byom}. 
Despite these promising developments, merging models that were fine-tuned with low-rank adaptation (LoRA) \citep{lora}, remains challenging and can severely degrade performance \citep{knots,llora}.

We argue that this degradation stems from parameter interference and how each model’s parameters interact with out-of-task data.
While prior work has focused on preserving orthogonality among task vectors \citep{tangentta,ethos,taujp} or aligning them in a shared space \citep{knots}, these data-free strategies often overlook the crucial interplay between latent features and parameter updates. 
Specifically, consider a pre-trained layer $W_0$ and two sets of learned LoRA blocks $\{B_1, A_1\}$ and $\{B_2, A_2\}$. 
The merged model is then formulated as $W_m = W_0 + B_1A_1+B_2A_2$.
Given a latent feature vector $\mathbf{h}_1$ from task $T_1$, the merged model produces $W_m\mathbf{h}_1=W_1\mathbf{h}_1+B_2A_2\mathbf{h}_1$, where the first term represents the intended response while the second term is the undesired shift.
Notably, existing data-free approaches that focus solely on resolving parameter conflicts are insufficient to mitigate such interference.

\begin{figure}[t]
    \centering
    \includegraphics[width=1.\linewidth]{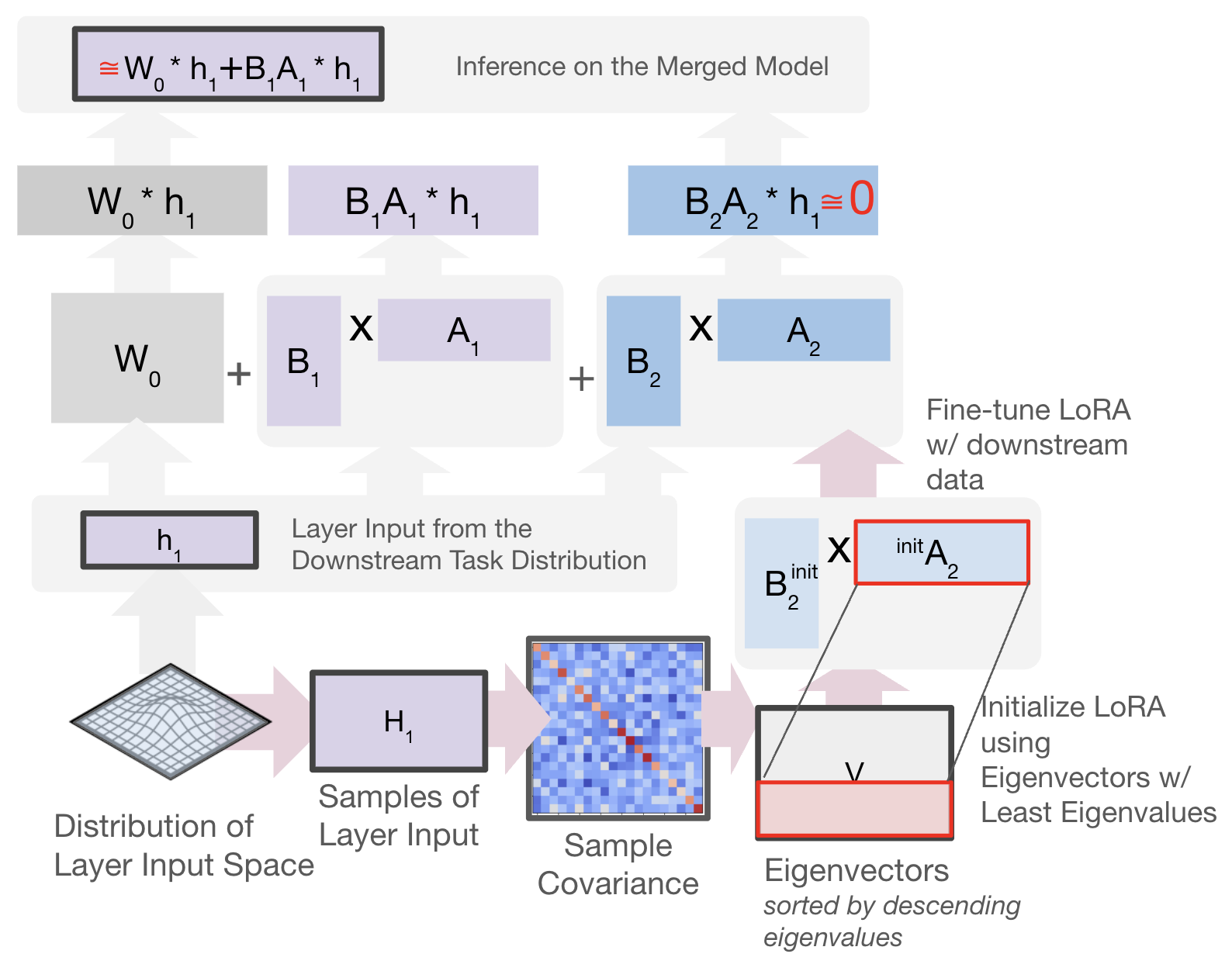}
    \caption{
    Overview of OSRM, which 
    seeks a data-driven subspace to initiate LoRA fine-tuning and thereby 
    greatly improves model performance when merging multiple LoRA
    models from different tasks.
    $W_0$ is the pre-trained weight.
    $\{B_i,A_i\}$ are LoRA fine-tuned on the $i$-th task.
    \textcolor[HTML]{c5b4f0}{Purple}: $(W_0+B_1A_1)*h_1$ is the required output.
    \textcolor[HTML]{93bff1}{Light blue}: Decompose the sample covariance matrix to initialize $A_2$.
    \textcolor[HTML]{267ad8}{Dark blue}: Reduce the output shift induced by $B_2A_2$.
    }
    \label{fig:paradigm}
    \vspace{-0.2in}
\end{figure}

In this paper, we propose a novel approach named \textbf{OSRM} (\textbf{O}rthogonal \textbf{S}ubspaces for \textbf{R}obust model \textbf{M}erging) that restricts the LoRA subspace \emph{before} fine-tuning, making it largely orthogonal to irrelevant out-of-task data distributions. 
Concretely, we aim to reduce the interference between data and parameters by minimizing $\|A_2\mathbf{h}_1\|_F$ and derive an analytical solution under the assumption that $A_2$ has an orthogonal basis.
Our method integrates seamlessly with existing merging algorithms and mitigates the unintentional output shifts that arise when multiple LoRA modules are combined (c.f.~\cref{fig:paradigm}). 
We further present practical extensions to ensure robust performance in real-world scenarios.

Extensive experiments on eight datasets using three widely used LMs and two large LMs confirm that our approach consistently outperforms existing merging baselines on multi-task evaluations, while preserving strong single-task performance. 
Additionally, our empirical results show the robustness of our method against hyperparameters, such as the scaling coefficient, the sample size, the number of tasks, and the choice of learnable blocks.
Our findings highlight the importance of considering data-parameter interplay in model merging and demonstrate a generalizable strategy for combining LoRA models more effectively.

\section{Related Work}
\label{sec:related work}

\paragraph{Model Merging.}  
A significant body of work explores merging models that have been fine-tuned independently on different datasets to obtain a unified multi-task model.  
\citet{taskarithmetic} introduced Task Arithmetic (TA), which defines task vectors for models and combines them using a unified weight.  
Building on TA, several methods have been proposed to address weight entanglement in task-specific models by aligning them before merging~\citep{ethos,tangentta,taujp,mats,knots,ta-vocabmodel,gitrebasin}.  
Other approaches improve TA by designing better weighting schemes for model averaging~\citep{fisher,regmean,lines,metagpt}.  
A different research direction focuses on manipulating task vectors to enhance merging performance~\citep{ties,pcb,byom,localizeandstitch,localizing,breadcrumbs}.  
Alternatively, some methods perform adaptive model merging using dynamic inference, employing either a router~\citep{twin,zipit} or masks~\citep{emr}.  

\paragraph{Merging LoRA Models.}  
LoRA~\citep{lora} has become a widely used technique for parameter-efficient fine-tuning.  
However, most existing model merging methods struggle to effectively transfer to LoRA models~\citep{knots,llora}.  
KnOTS~\citep{knots} highlights the importance of merging LoRA models within a shared space, while \citet{llora} propose that linearized LoRA models reduce weight entanglement.  
\citet{lego-lora} and \citet{lora-soups} further improve LoRA merging by learning optimal merging weights and clustering LoRA modules, respectively.  

Despite these advancements, existing methods often overlook the interaction between model parameters and data, leading to suboptimal merging performance.  
Moreover, prior work primarily focuses on the phases \emph{during} and \emph{after} fine-tuning.  
In contrast, our method explicitly addresses parameter-data interactions and focuses on the merging phase \emph{before} fine-tuning.

\section{Preliminaries and Background}
\label{sec:pre}

\paragraph{Notation.}
Let $f_0$ be a pre-trained model with $L$ layers where $\theta_0=\{W_0^{(1)},\dots,W_0^{(L)}\}$ is the parameters of the model and $W_0^{(l)}$ is the weight matrix of the $l$-th layer. 
During adaptation, $f_0$ is individually fine-tuned on $N$ downstream tasks $\{T_1,\dots,T_N\}$ with $\theta_t$ the fine-tuned parameters on task $T_t$ where $\theta_t = \{W_t^{(1)},\dots,W_t^{(L)}\}$.

\paragraph{Low-Rank Adaptation.}
Low-rank adaptation (LoRA)~\cite{lora} is a popular method for efficiently fine-tuning a pre-trained model on a downstream task.
It introduces low-rank subspaces to contain the parameter updates during fine-tuning.
Specifically, the parameter update for a weight matrix $W \in \mathbb{R}^{m \times n}$ is represented as $\Delta W = BA$, where $B \in \mathbb{R}^{m \times r}$ and $A \in \mathbb{R}^{r \times n}$ are two learnable matrices with $r \ll \min(m, n)$.
Typically, $B$ and $A$ are initialized as zeros and random Gaussian noise before fine-tuning, respectively, and then learned during the fine-tuning phase.
We denote the latent feature at the $l$-th layer as: 
\begin{align*}
    \mathbf{h}^{(l)} 
    &= W_0^{(l)}\mathbf{h}^{(l-1)} + \Delta W^{(l)}\mathbf{h}^{(l-1)} \notag \\
    &= W_0^{(l)}\mathbf{h}^{(l-1)} + B^{(l)}A^{(l)}\mathbf{h}^{(l-1)}.
\end{align*}

\paragraph{Model Merging.}
\label{para:model merging}
We focus on two categories of model merging: weighted averaging and task vector manipulations.

\noindent\emph{Task Arithmetic} (TA)~\citep{taskarithmetic} merges models by linearly summing their parameters as $\theta_0 + \lambda \sum_{t} (\theta_t-\theta_0)$, where $\lambda$ is a scaling coefficient tuned on a validation set. 
\emph{Fisher Merging}~\citep{fisher} improves upon TA by modeling each model's posterior as a Gaussian distribution, leveraging Fisher information to weight model averages instead of using a single uniform coefficient.
\emph{RegMean}~\citep{regmean} extends model merging by drawing inspiration from linear models, aiming to minimize transformation shifts on data before and after merging. 
The design leads to an analytical solution that generalizes to language models.

\emph{TIES}~\citep{ties} further refines TA by operating at the level of task vectors. It first reduces redundancy by pruning low-magnitude parameters, then resolves parameter conflicts by selecting dominant signs, and finally merges only the aligned parameters.
For more flexible merging, \citet{emr} propose an adaptive approach \emph{EMR} that generates a task-specific model at inference time. Their method selects a unified base model from a set of candidates and dynamically applies task-specific masks and rescaling factors.

\section{Our Proposed Method}
\label{sec:method}

In this section, we first introduce our proposed method OSRM (\textbf{O}rthogonal \textbf{S}ubspaces for \textbf{R}obust model \textbf{M}erging) for constraining the transformation capacity of the LoRA subspace \emph{before} fine-tuning to improve the performance of model merging \emph{after} fine-tuning.
We then propose several practical extensions to facilitate the integration of our method into real-world scenarios.

\subsection{Motivation}
\label{subsec:motivation}

Existing approaches often seek to eliminate interference among multiple models via weight disentanglement, such as orthogonalizing task vectors.
Recently, \citet{knots} argued that task-vector orthogonality does not necessarily imply no interference and proposed a data-free method to align LoRA modules in a shared space.
Despite its promising empirical results, data-free approaches ignore how parameters interact with the input features in each layer and may be suboptimal in effectively mitigating interference.

Without loss of generality, let us consider merging two tasks $T_1$ and $T_2$ via task arithmetic as an illustrative example; we omit the subscript $l$ for the layer index.
The merged weight matrix is:
\begin{equation*}
    W_m 
    = W_0 + \Delta W_1 + \Delta W_2 
    = W_0 + B_1 A_1 + B_2 A_2,
\end{equation*}
where $\{B_t, A_t\}$ denote the LoRA matrices for task~$t$.
During inference, given a latent feature $\mathbf{h}_1$ that arises from a sample of task~$T_1$, the transformation of $W_m$ on~$\mathbf{h}_1$ is:
\begin{align}
    W_m \mathbf{h}_1
    &= (W_0 + B_1 A_1 + B_2 A_2) \mathbf{h}_1 \notag \\
    &= (W_0 + B_1 A_1) \mathbf{h}_1 + B_2 A_2 \mathbf{h}_1 \notag \\
    &= W_1 \mathbf{h}_1 + B_2 A_2\,\mathbf{h}_1,
    \label{eq:interference}
\end{align}
where $W_1 = W_0 + B_1 A_1$ is the fine-tuned model for task~$T_1$, and $B_2 A_2 \,\mathbf{h}_1$ can be viewed as a ``perturbation'' induced by the knowledge learned for task~$T_2$.
To reduce the influence of $B_2 A_2$ on $\mathbf{h}_1$, we must consider this interaction and limit the transformation capacity of $B_2 A_2$ when operating on $\mathbf{h}_1$.
This motivated us to propose a novel approach. 

\subsection{Constraining the LoRA Subspace}
\label{subsec:objective}

\begin{figure}[t]
    \centering
    \includegraphics[width=0.8\linewidth]{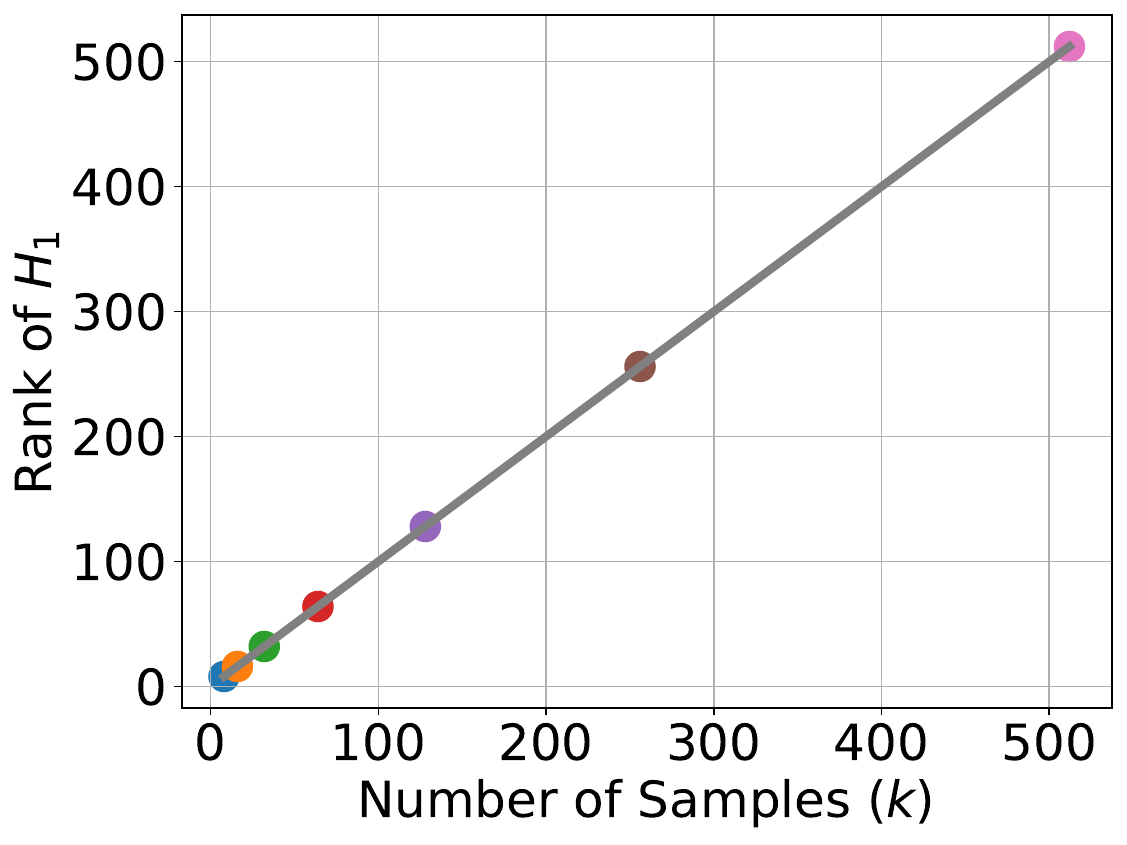}
    \caption{The rank of $H_1$ (y-axis) vs. the number of samples $k$ (x-axis) with RoBERTa-large~\citep{roberta}. The grey line represents $y=x$. For each dot, $k$ samples are randomly selected to concatenate their latent features as $H_1$.}
    \label{fig:rank of H1}
\end{figure}

We first generalize \cref{eq:interference} to the situation where $k>1$ samples (from task~$T_1$) are available to characterize the interference.
Let $H_1 \in \mathbb{R}^{k \times n}$ be the matrix whose rows are the latent features of these $k$ samples. 
The merged transformation becomes:
\[
    W_m H_1^\top = W_1 H_1^\top + B_2 A_2 H_1^\top.
\]
To make $W_m H_1^\top$ as close as possible to $W_1 H_1^\top$, ideally one wants to force $A_2 H_1^\top = \mathbf{0}$.
If $H_1$ is rank-deficient, we could simply constrain each row of $A_2$ to be in the null space of $H_1$. 
However, \cref{fig:rank of H1} shows that $H_1$ is typically full-rank in real scenarios due to intrinsic data variability, so we cannot generally make $A_2 H_1^\top = \mathbf{0}$.
Instead, we propose to \emph{reduce} the interference by minimizing its Frobenius norm:
\[
    \min_{A_2} \; \bigl\|A_2 H_1^\top\bigr\|_F.
\]
We further opt \emph{not} to constrain $B_2$ because stringent constraints on both matrices could degrade the representation power needed for task~$T_2$.
Thus, our strategy is to disentangle the LoRA matrices so that $A_2$ is constrained \emph{prior to} fine-tuning to reduce interference with other tasks, while $B_2$ remains free to maintain downstream performance.

Directly minimizing $\|A_2 H_1^\top\|_F$ without constraints would trivially yield $A_2=\mathbf{0}$.
In practice, although $\{B, A\}$ serve as low-rank approximations of the fine-tuned weight $\Delta W$, the learned $\{B, A\}$ are mostly full-rank.
Because any full-rank matrices can be factorized using RQ decomposition, we can safely set $A$ to be an orthogonal basis.

To balance these considerations, we require $A_2$ to have orthonormal rows, i.e.\ $A_2 A_2^\top = I$. 
This enforces a full-rank condition on the row space of $A_2$ while removing the scale ambiguity between $B_2$ and $A_2$. 
Indeed, for any non-zero scalar~$c$, $(c B_2)\,(\tfrac{1}{c} A_2)$ gives the same product as $B_2 A_2$.
Imposing $A_2 A_2^\top = I$ fixes the scaling of $A_2$ and forces $B_2$ to handle the appropriate scaling.
To this end, we arrive at the following optimization:
\begin{equation}
    \Tilde{A_2} 
    \;=\; \argmin\nolimits_{A}\; \bigl\| A\,H_1^\top \bigr\|_F^2,
    \  \st \ 
    A\,A^\top = I.
    \label{eq:objective}
\end{equation}
We show that it admits an analytical solution.

\subsection{Analytical Solution}
\label{subsec:solution}

For brevity, we temporarily drop the subscripts (i.e.\ write $A$ instead of $A_2$, and $H$ instead of $H_1$).
Let $S = \frac{1}{k-1}H^\top H$ be the sample covariance matrix of~$H$. Since $S$ is symmetric and positive semi-definite, we can perform the eigendecomposition:
\[
    S = V \Lambda V^\top,
\]
where $V \in \mathbb{R}^{n \times n}$ is an orthogonal matrix and 
$\Lambda = \diag(\lambda_1,\dots,\lambda_n)$ contains the eigenvalues.
With a descending order of eigenvalues $\lambda_1 \ge \lambda_2 \ge \cdots \ge \lambda_n$, the analytical solution to~\cref{eq:objective} is:
\begin{align}
    \label{eq:solution}
    \Tilde{A_2} = 
        V^\top_{:\,,\,n-r:n},
\end{align}
where $V_{:\,,\,n-r:n}$ denotes the last $r$ eigenvectors of $S$ corresponding to the $r$ smallest eigenvalues.
See~\cref{app:proof} for detailed derivation.
Although the theory constrains $A_2$, in practice, one might still update $A_2$ to prevent excessive loss of accuracy on the target task (c.f. \cref{subsec:extension}).

\paragraph{Interpretation.}
Intuitively, $\|A_2 H_1^\top\|_F^2 = \mathrm{tr}(A_2 S A_2^\top)$ measures how $A_2$ amplifies the principal directions of $H_1$ in its row space.
By choosing directions corresponding to the smallest eigenvalues of $S$, we place $A_2$ in the subspace where $H_1$ has the minimal variance, thereby reducing the interference term $B_2 A_2 \,H_1^\top$ most effectively.
Meanwhile, the row-orthonormal constraint $A_2 A_2^\top = I$ preserves non-trivial capacity for $B_2$ to learn scaling factors and ensure the model can still fit task~$T_2$.

\begin{algorithm}[t]
    \caption{Model merging with OSRM}
    \label{alg:method}
      \begin{flushleft}
        \textbf{Input:} a pre-trained model $f_0$, tasks $[T_1,\dots,T_N]$, a merging method $\mathcal{M}$, number of layers $L$.
        \end{flushleft}
   \begin{algorithmic}[1]
        \STATE \textit{/* Constrain the LoRA subspace */}
        \FOR{$t=1,\cdots,N$}
            \STATE Randomly select $k$ validation samples from $T_t$
            \STATE Collect and concatenate latent features $\{H_t\}_{l=1}^L$ for all layers
            \STATE Average sample-wise features to get $\{\bar{H}_{t}^{(l)}\}_{l=1}^L$ based on \cref{eq:feature concat}
        \ENDFOR
        \STATE \textit{/* Fine-tune on downstream tasks */}
        \FOR{$t=1,\cdots,N$}
            \STATE Initialize $\{A_t^{(l)}\}_{l=1}^L$ to the solution in \cref{eq:solution}
            \STATE Initialize $\{B_t^{(l)}\}_{l=1}^L$ to zeros
            \STATE Fine-tune $\{B_t^{(l)}, A_t^{(l)}\}_{l=1}^L$ on $T_t$ to get $\theta_t$
        \ENDFOR
        \STATE \textit{/* Merge fine-tuned models */}
        \STATE Merge the fine-tuned models with an existing method $\theta_m = \mathcal{M}(\theta_1,\cdots,\theta_N)$
   \end{algorithmic}
\end{algorithm}

\begin{figure}[t]
    \centering
    \includegraphics[width=0.85\linewidth]{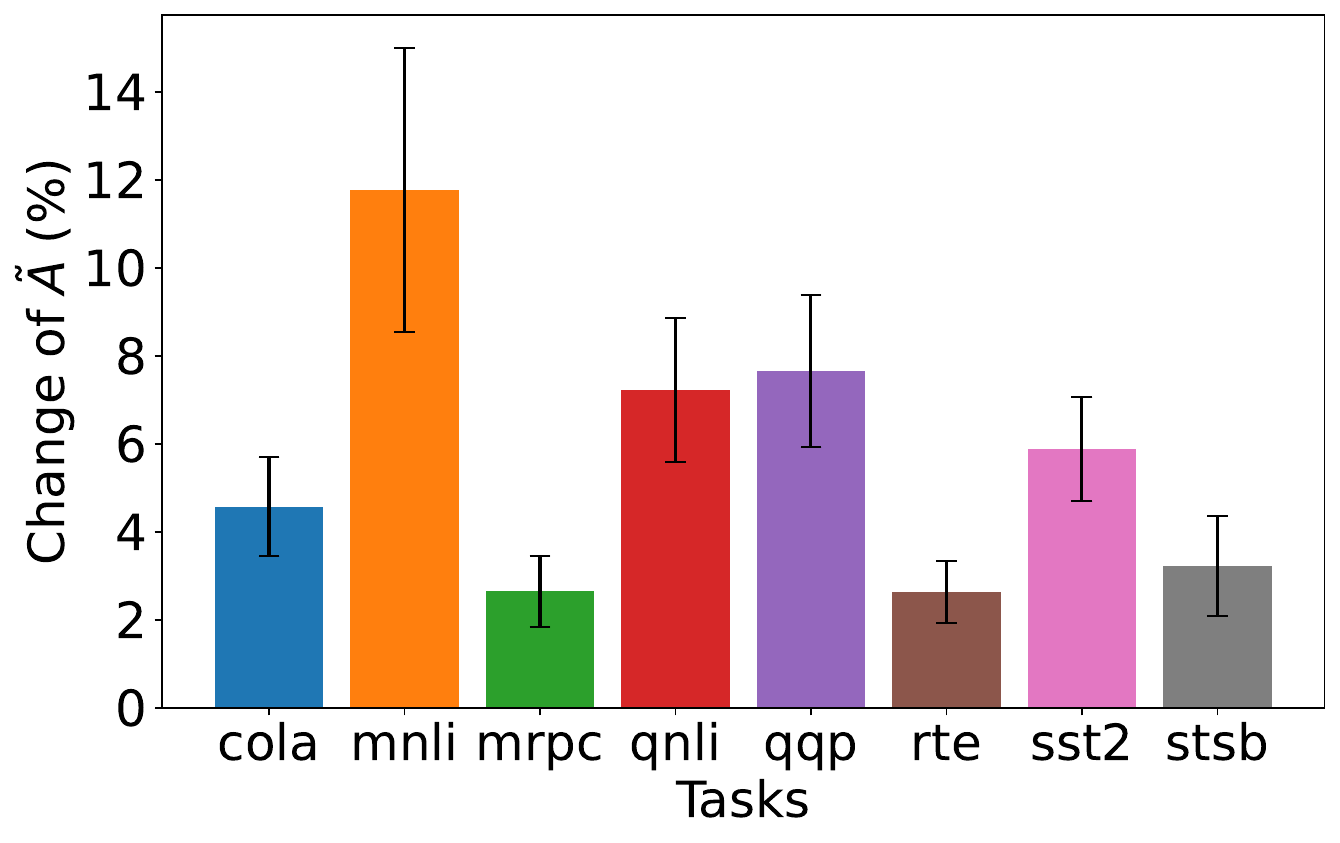}
    \caption{The change of $\Tilde{A}$ ($\%$) after fine-tuning compared to the initialization. A normalized distance is used as the metric. See~\cref{subsec:extension} for details.}
    \label{fig:change of A}
\end{figure}

\subsection{Practical Extensions}
\label{subsec:extension}

We now discuss practical extensions that facilitate the adaptation of OSRM to real-world applications.
\Cref{alg:method} summarizes the overall procedure to merge models with OSRM.

\paragraph{Relaxing the Constraint during Fine-Tuning.}
To minimize interference among merged models, our analysis suggests freezing $\Tilde{A_2}$ at its solution in \cref{eq:solution} during fine-tuning. 
Nonetheless, our empirical results show this can significantly degrade single-task accuracy.
One possible reason for the performance drop is that fixing $A_2$ restricts the model's adaptation capability and hurts performance on $T_2$.
To avoid this, we propose that $\Tilde{A_2}$ should only be used as the \emph{initialization} of $A_2$, and allow it to be updated during fine-tuning, which we show significantly improves single-task performance.

To show that $\Tilde{A}_2$ is still validly orthogonal to the latent features, we adopt the orthogonal Procrustes problem~\citep{procrustes}:
\[
    D=\min\nolimits_{\Omega} \| \Omega \Tilde{A}^{\textrm{ft}} - \Tilde{A}^{\textrm{init}} \|_F,\;\st\; \Omega^\top \Omega = I,
\]
where $\Tilde{A}^{\textrm{ft}}$ and $\Tilde{A}^{\textrm{init}}$ are the fine-tuned and initialized matrices of $\Tilde{A}$, respectively.
Thus, $D$ measures the distance between two matrices under orthogonal transformations.
We then use the normalized distance as the metric to measure the change of $\Tilde{A}^{\textrm{init}}$ after fine-tuning, denoted as $D / \| \Tilde{A}^{\textrm{init}} \|_F$.
Results in~\cref{fig:change of A} show that the change of $\Tilde{A}$ is up to $14\%$ approximately, which is marginal, implying the validity of our relaxation.

\paragraph{Extension to Multiple Tasks.}
When merging more than two tasks, say $T_1,\dots,T_N$, one can gather latent features from all tasks \emph{except} $T_t$ to build $\Tilde{A_t}$.
Concretely, let $H_i$ be the latent feature matrix of task~$T_i$.
For task $T_t$, we concatenate the features of the other tasks into
\begin{align*}
    H_{\neg t} = \bigl[H_1;\cdots;H_{t-1};H_{t+1};\cdots;H_N\bigr].
\end{align*}
Then the objective in \cref{eq:objective} becomes:
\begin{align*}
        \Tilde{A_t} 
        =
        \argmin\nolimits_{A} \bigl\|A\,H_{\neg t}^\top\bigr\|_F^2
        \quad \st \quad
        A\,A^\top=I.
\end{align*}
In large-scale or privacy-sensitive applications, storing or sharing \emph{all} latent features can be memory-intensive or prohibitive.
A practical fix is to \emph{average} the sample-wise latent features in each task:
\begin{align}
    \bar{H}_i = \frac{1}{k} \sum\nolimits_{j=1}^{k} H_{i;j,:},
    \label{eq:feature concat}
\end{align}
where $H_{i;j,:}$ is the $j$-th row of $H_i$. 
We then replace $H_{\neg t}$ with the concatenation of $\{\bar{H}_i \mid i \ne t\}$ to reduce memory usage and mitigate privacy concerns.

\section{Experiments}
\label{sec:experiments}

\begin{table*}[t]
    \small
    \centering
    \caption{Per-task performance ($\%$) of merging fine-tuned RoBERTa-large models.
    "Individual" refers to the performance of each fine-tuned model on the dataset on which it was trained.
    \textbf{Bold} indicates a better performance.
    }
    \label{tab:main-roberta-pertask}
    \resizebox{.95\textwidth}{!}{%
    \begin{tabular}{cc|cccccccc|
    >{\columncolor[HTML]{EFEFEF}}c }
    \hline
     & OSRM & CoLA & MNLI & MRPC & QNLI & QQP & RTE & SST2 & STSB & Avg. \\ \hline
     & No & \textbf{68.75} & \textbf{89.05} & 91.67 & 94.23 & 89.38 & 83.03 & \textbf{96.33} & 91.92 & \textbf{88.05} \\
    \multirow{-2}{*}{Individual} & Yes & 66.28 & 88.88 & 91.67 & \textbf{94.62} & \textbf{89.45} & \textbf{83.75} & 96.1 & \textbf{92.17} & 87.87 \\ \hline
     & No & 18.57 & 74.01 & 77.7 & 73.75 & 82.84 & 71.12 & 87.16 & 75.13 & 70.04 \\
    \multirow{-2}{*}{TA} & Yes & \textbf{32.25} & \textbf{81.24} & 77.7 & \textbf{86.78} & \textbf{84.99} & \textbf{80.14} & \textbf{91.63} & \textbf{77.96} & \textbf{76.59} \\ \hline
     & No & 31.77 & 66.85 & 69.61 & \textbf{87.42} & \textbf{83.94} & \textbf{72.56} & \textbf{94.72} & 36.19 & 67.88 \\
    \multirow{-2}{*}{RegMean} & Yes & \textbf{40.36} & \textbf{71.45} & \textbf{76.96} & 72.01 & 76.59 & 63.54 & 92.32 & \textbf{65.28} & \textbf{69.81} \\ \hline
     & No & 36.16 & 53.62 & 70.1 & \textbf{65.15} & 71.23 & 67.15 & 91.17 & 64.45 & 64.88 \\
    \multirow{-2}{*}{Fisher} & Yes & \textbf{39.11} & \textbf{74.85} & \textbf{76.72} & 62.51 & \textbf{81.64} & 67.15 & \textbf{93.35} & \textbf{79.56} & \textbf{71.86} \\ \hline
     & No & 13.06 & 70.23 & \textbf{63.73} & \textbf{83.64} & 83.36 & 63.18 & 87.61 & 25.75 & 61.32 \\
    \multirow{-2}{*}{TIES} & Yes & \textbf{29.88} & \textbf{77} & 58.24 & 83.52 & \textbf{84.39} & \textbf{72.92} & \textbf{90.48} & \textbf{36.75} & \textbf{66.65} \\ \hline
     & No & 54.69 & 83.31 & \textbf{79.66} & 89.35 & 85.04 & \textbf{81.95} & 92.78 & 82.63 & 81.18 \\
    \multirow{-2}{*}{EMR} & Yes & \textbf{56.22} & \textbf{88.51} & 77.21 & \textbf{93.3} & \textbf{87.54} & 79.78 & \textbf{95.41} & \textbf{88.56} & \textbf{83.32} \\ \hline
    \end{tabular}%
    }
\end{table*}

\begin{table*}[t]
    \small
    \centering
    \caption{Per-task performance (\%) of T5-large. "Individual" refers to the metrics of each fine-tuned model on the dataset on which it was trained. \textbf{Bold} indicates a better performance.}
    \label{tab:t5-pertask}
    \resizebox{.95\textwidth}{!}{%
    \begin{tabular}{cc|cccccccc|
    >{\columncolor[HTML]{EFEFEF}}c }
    \hline
     & OSRM & CoLA & MNLI & MRPC & QNLI & QQP & RTE & SST2 & STSB & Avg. \\ \hline
     & No & 60.86 & 88.17 & 88.73 & 94.25 & 90.42 & 81.59 & 95.76 & 91.25 & 86.38 \\
    \multirow{-2}{*}{Individual} & Yes & \textbf{62.58} & \textbf{88.31} & \textbf{90.69} & \textbf{94.62} & \textbf{91.04} & 81.59 & \textbf{95.87} & \textbf{91.36} & \textbf{87.01} \\ \hline
     & No & \textbf{28.04} & 68.10 & \textbf{44.85} & 56.07 & \textbf{35.35} & 58.48 & 56.08 & 76.74 & 52.96 \\
    \multirow{-2}{*}{TA} & Yes & 25.24 & \textbf{70.83} & 28.92 & \textbf{79.21} & 34.50 & \textbf{65.70} & \textbf{62.16} & \textbf{79.79} & \textbf{55.79} \\ \hline
     & No & 41.22 & \textbf{67.92} & 53.68 & 84.77 & 87.02 & 68.23 & \textbf{93.00} & 63.51 & 69.92 \\
    \multirow{-2}{*}{RegMean} & Yes & \textbf{43.67} & 66.19 & \textbf{69.36} & \textbf{87.24} & \textbf{88.32} & \textbf{70.76} & 92.89 & \textbf{70.98} & \textbf{73.68} \\ \hline
     & No & 13.72 & \textbf{35.45} & 31.62 & 49.46 & 63.18 & \textbf{52.71} & 49.08 & \textbf{90.60} & 48.23 \\
    \multirow{-2}{*}{Fisher} & Yes & \textbf{38.91} & 34.68 & \textbf{68.38} & \textbf{51.22} & 63.18 & 49.10 & \textbf{53.56} & 90.02 & \textbf{56.13} \\ \hline
     & No & 36.67 & \textbf{73.13} & 29.90 & 62.46 & 39.78 & 64.26 & 63.65 & \textbf{74.07} & 55.49 \\
    \multirow{-2}{*}{TIES} & Yes & \textbf{40.16} & 70.27 & \textbf{33.09} & \textbf{82.34} & \textbf{48.51} & \textbf{64.62} & \textbf{65.37} & 73.57 & \textbf{59.74} \\ \hline
     & No & 37.16 & 87.27 & 79.66 & 93.30 & 88.32 & \textbf{80.14} & 94.27 & 88.08 & 81.02 \\
    \multirow{-2}{*}{EMR} & Yes & \textbf{39.54} & \textbf{88.40} & \textbf{85.78} & \textbf{94.07} & \textbf{89.52} & 79.90 & \textbf{94.95} & \textbf{88.72} & \textbf{82.61} \\ \hline
    \end{tabular}%
}
\end{table*}

\begin{table*}[t]
    \small
    \centering
    \caption{Per-task performance ($\%$) of Llama3.2-1B. "Individual" refers to the metrics of each fine-tuned model on the dataset on which it was trained. \textbf{Bold} indicates a better performance.}
    \label{tab:llama1b-pertask}
    \resizebox{.87\textwidth}{!}{%
    \begin{tabular}{cc|cccccccc|
    >{\columncolor[HTML]{EFEFEF}}c }
    \hline
     & OSRM & CoLA & MNLI & MRPC & QNLI & QQP & RTE & SST2 & STSB & Avg. \\ \hline
     & No & 59.97 & 85.44 & \textbf{87.25} & 91.10 & 88.34 & 78.34 & 94.84 & \textbf{90.10} & 84.42 \\
    \multirow{-2}{*}{Individual} & Yes & \textbf{61.44} & \textbf{86.91} & 86.52 & \textbf{92.26} & \textbf{89.07} & \textbf{82.67} & \textbf{95.76} & 89.56 & \textbf{85.52} \\ \hline
     & No & 26.05 & \textbf{72.63} & 66.91 & 60.70 & \textbf{82.61} & 57.04 & \textbf{92.32} & 67.15 & 65.68 \\
    \multirow{-2}{*}{TA} & Yes & \textbf{28.20} & 66.69 & 66.91 & \textbf{73.48} & 81.66 & \textbf{68.23} & 92.20 & \textbf{71.24} & \textbf{68.58} \\ \hline
     & No & 30.55 & \textbf{34.08} & 41.42 & 48.89 & 52.48 & 50.54 & 48.28 & 23.31 & 41.19 \\
    \multirow{-2}{*}{RegMean} & Yes & \textbf{34.96} & 32.97 & \textbf{43.14} & \textbf{49.99} & \textbf{60.39} & \textbf{53.07} & \textbf{48.62} & \textbf{25.45} & \textbf{43.57} \\ \hline
     & No & \textbf{19.91} & 60.47 & \textbf{69.85} & \textbf{60.28} & \textbf{82.18} & 70.40 & 92.09 & \textbf{33.29} & \textbf{61.06} \\
    \multirow{-2}{*}{Fisher} & Yes & 19.40 & \textbf{62.42} & 69.61 & 59.93 & 74.86 & \textbf{71.99} & \textbf{92.34} & 26.58 & 59.64 \\ \hline
     & No & 41.18 & 81.77 & 76.47 & \textbf{87.35} & 85.08 & \textbf{75.45} & 92.43 & 83.04 & 77.85 \\
    \multirow{-2}{*}{TIES} & Yes & \textbf{42.02} & \textbf{83.03} & \textbf{78.43} & 87.04 & \textbf{87.35} & 74.73 & \textbf{94.95} & \textbf{82.62} & \textbf{78.77} \\ \hline
     & No & 41.18 & 81.77 & 80.69 & \textbf{87.35} & 83.05 & \textbf{75.45} & 92.43 & \textbf{83.04} & 78.12 \\
    \multirow{-2}{*}{EMR} & Yes & \textbf{42.02} & \textbf{83.03} & \textbf{81.44} & 87.04 & \textbf{84.97} & 74.73 & \textbf{94.95} & 82.62 & \textbf{78.85} \\ \hline
    \end{tabular}%
    }
\end{table*}

\begin{table*}[t]
    \small
    \centering
    \caption{Per-task performance ($\%$) of Llama3.2-3B. "Individual" refers to the metrics of each fine-tuned model on the dataset on which it was trained. \textbf{Bold} indicates a better performance.}
    \label{tab:llama3b-pertask}
    \resizebox{.87\textwidth}{!}{%
    \begin{tabular}{cc|cccccccc|
    >{\columncolor[HTML]{EFEFEF}}c }
    \hline
     & OSRM & CoLA & MNLI & MRPC & QNLI & QQP & RTE & SST2 & STSB & Avg. \\ \hline
     & No & 68.61 & 89.31 & \textbf{87.01} & 93.26 & 89.81 & 85.92 & 96.10 & \textbf{90.25} & 87.54 \\
    \multirow{-2}{*}{Individual} & Yes & \textbf{69.25} & \textbf{89.57} & 86.52 & \textbf{94.51} & \textbf{90.15} & \textbf{89.89} & \textbf{96.90} & 89.82 & \textbf{88.33} \\ \hline
     & No & 34.80 & 77.87 & 72.30 & 80.03 & 84.64 & \textbf{68.23} & \textbf{93.46} & 74.47 & 73.22 \\
    \multirow{-2}{*}{TA} & Yes & \textbf{35.50} & \textbf{83.79} & \textbf{72.79} & \textbf{84.50} & \textbf{84.79} & 63.54 & 92.78 & \textbf{77.20} & \textbf{74.36} \\ \hline
     & No & 40.61 & 33.46 & \textbf{50.74} & 50.03 & 50.84 & 46.21 & 50.46 & \textbf{41.90} & 45.53 \\
    \multirow{-2}{*}{RegMean} & Yes & \textbf{44.04} & \textbf{33.51} & 42.16 & \textbf{50.72} & \textbf{53.72} & \textbf{49.82} & \textbf{51.11} & 41.10 & \textbf{45.77} \\ \hline
     & No & \textbf{28.21} & 34.14 & 62.50 & 66.17 & 73.50 & \textbf{54.15} & 89.91 & 29.24 & 54.73 \\
    \multirow{-2}{*}{TIES} & Yes & 22.75 & \textbf{41.06} & \textbf{65.20} & \textbf{71.41} & \textbf{75.88} & 49.82 & \textbf{90.02} & \textbf{42.12} & \textbf{57.28} \\ \hline
     & No & 54.74 & 87.92 & \textbf{84.31} & 91.34 & 87.52 & \textbf{84.12} & 95.64 & 81.86 & 83.43 \\
    \multirow{-2}{*}{EMR} & Yes & \textbf{55.57} & \textbf{89.43} & 80.15 & \textbf{93.61} & \textbf{88.72} & 79.42 & \textbf{96.56} & \textbf{84.32} & \textbf{83.47} \\ \hline
    \end{tabular}%
    }
\end{table*}

\begin{table*}[t]
\centering
\caption{Per-task performance (\%) of Llama3-8B. "Individual" refers to the metrics of each fine-tuned model on the dataset on which it was trained.
\textbf{Bold} indicates a better performance.}
\label{tab:llama8b-pertask}
\resizebox{.8\textwidth}{!}{%
\begin{tabular}{cc|cccccccc
>{\columncolor[HTML]{EFEFEF}}c}
\hline
 & OSRM & CoLA & MNLI & MRPC & QNLI & QQP & RTE & SST2 & STSB & Avg. \\ \hline
 & No & 67.57 & 88.74 & \textbf{89.95} & 94.65 & \textbf{90.26} & 84.48 & 96.44 & \textbf{91.40} & 87.94 \\
\multirow{-2}{*}{Individual} & Yes & \textbf{69.20} & \textbf{90.15} & 89.46 & \textbf{95.44} & 90.17 & \textbf{88.09} & \textbf{97.02} & 90.65 & \textbf{88.77} \\ \hline
 & No & 10.00 & 64.74 & 31.62 & 64.74 & \textbf{81.22} & 80.87 & \textbf{92.89} & \textbf{80.14} & 63.28 \\
\multirow{-2}{*}{TA} & Yes & \textbf{52.89} & \textbf{82.53} & 31.62 & \textbf{89.58} & 72.56 & \textbf{87.00} & 88.76 & 77.08 & \textbf{72.75} \\ \hline
 & No & \textbf{28.28} & \textbf{35.68} & \textbf{41.18} & 65.62 & 66.97 & \textbf{61.73} & \textbf{78.67} & \textbf{45.97} & \textbf{53.01} \\
\multirow{-2}{*}{TIES} & Yes & 27.75 & 33.54 & 36.52 & \textbf{71.19} & \textbf{77.10} & 59.46 & 74.56 & 41.36 & 52.69 \\ \hline
\end{tabular}
}
\end{table*}

\subsection{Experimental Settings}
\label{subsec:experiment setting}

\paragraph{Datasets.}
We evaluate our method using eight datasets from the GLUE benchmark~\citep{glue}, a widely used suite of natural language understanding tasks. 
These tasks encompass both single-sentence and sentence-pair classification, including MRPC~\citep{mrpc}, QQP~\citep{qqp}, QNLI~\citep{qnli}, MNLI~\citep{mnli}, SST-2~\citep{sst2}, CoLA~\citep{cola}, STS-B~\citep{stsb}, and RTE~\citep{rte}. 
Each dataset is split into training, validation, and test sets. 

For evaluation, we use the Matthews correlation coeff. for CoLA and the average of the Pearson and Spearman correlation coeff. for STS-B, and accuracy is used for the remaining tasks. 
Since our method is applied \emph{before} fine-tuning and influences the training, we report absolute metric values in all experiments rather than normalized scores.

\paragraph{Models.}
To assess the effectiveness of our approach across different architectures, we evaluate three language models: 
RoBERTa-large~\citep{roberta} (encoder-only), T5-large~\citep{t5} (encoder-decoder), and Llama3.2-1B~\citep{llama3} (decoder-only). 
Additionally, we test our method on the larger Llama3.2-3B and Llama3-8B~\citep{llama3}. The main results for these models are presented in~\cref{subsec:main results}.

\paragraph{Baselines.}  
We evaluate the effectiveness of our method against five widely used model merging techniques.  
\citet{taskarithmetic} introduced Task Arithmetic (TA) to merge models with a unified weight. 
Fisher merging~\citep{fisher} and RegMean~\citep{regmean} improve upon TA by incorporating anisotropic weighting, utilizing Fisher information and the inner product of data matrices, respectively.  
TIES~\citep{ties} performs merging at the level of task vectors, while EMR~\citep{emr} represents the state-of-the-art in adaptive model merging.  
For additional background, refer to~\cref{para:model merging}.

\paragraph{Implementation Details.}
For training, we follow~\cite{lora} and use the AdamW optimizer~\citep{adamw} with a warmup ratio of $0.06$ and a linear learning rate schedule. 
Following~\citep{lora}, LoRA is configured with a rank of $r=8$, a scaling factor of $\alpha=16$, and is only applied to the query and value blocks. 
We study the effect of different learnable blocks in~\cref{subsec:robustness}.
Other hyperparameters are selected via grid search as in~\citep{roberta} (c.f.~\cref{append:hyperparam}).

For model merging, we adopt hyperparameter settings from prior work~\citep{ties,regmean,fisher}. 
Specifically, we set the scaling coefficient to $0.3$ for TA and $1$ for TIES. The non-diagonal multiplier in RegMean is set to $0.9$, except for T5-large, where it is $0.1$. 
For Fisher-based merging, we use a uniform scaling factor of $\frac{1}{8}$ across all models. 
For methods requiring validation data, we use up to $1000$ samples from the validation set, following~\citep{regmean}. 
We use $100$ samples per task in our method to compute $H_t$.
The effect of hyperparameter variations is analyzed in~\cref{subsec:robustness}.
Our code implementation is adapted from \citep{emr} and available at \url{https://github.com/illidanlab/OSRM}. 

\subsection{Main Results}
\label{subsec:main results}

\paragraph{Merging Encoder-Only Models.}
The performance of merging RoBERTa-large models is presented in~\cref{tab:main-roberta-pertask}. 
Our proposed method consistently outperforms all the baselines across all merging techniques on average. 

Specifically, for TA merging, our method achieves performance at least on par with those baselines. It surpasses the baselines in seven out of eight tasks, with a notable improvement of over $13\%$ on CoLA. 
In the Fisher merging setting, our method slightly underperforms on the QNLI dataset, with a marginal gap of less than $3\%$, but outperforms the baseline across all other tasks, achieving up to a $21\%$ improvement on MNLI. 
For TIES and EMR, while the baseline slightly surpasses our method on two datasets, our approach yields significantly better overall performance across the eight datasets. 

Moreover, our method minimally impacts downstream task performance, with an average performance gap of less than $1\%$, and even surpasses the baseline on four datasets.

\paragraph{Merging Encoder-Decoder Models.}
The results for T5-large are reported in~\cref{tab:t5-pertask}. 
Our method substantially outperforms the baseline on RegMean, Fisher, and TIES.
On average, it improves RegMean and Fisher by $3.76\%$ and $7.9\%$, respectively. 
Notably, our approach enhances performance on MRPC by approximately $16\%$ under RegMean and on QQP by around $9\%$ under TIES. 

Although the baseline slightly outperforms our method on Fisher merging, the difference is minimal at just $0.07\%$. 
Additionally, our method consistently improves downstream task performance across all eight datasets.

\paragraph{Merging Decoder-Only Models.}
The results for Llama3.2-1B are shown in~\cref{tab:llama1b-pertask}. 
On average, our method surpasses the baseline on TA, RegMean, TIES, and EMR.
Per-task improvements reach up to $12.78\%$ and $7.91\%$ for TA and RegMean, respectively. 

Importantly, our method consistently enhances downstream fine-tuning performance on Llama3.2-1B, demonstrating its effectiveness in decoder-only model merging.

\paragraph{Merging Large Language Models.}
Following the setting of baselines, we further evaluate the effectiveness of our method on the large language model Llama3.2-3B, as shown in~\cref{tab:llama3b-pertask}.  
Due to the large scale of Llama3.2-3B, we focus solely on gradient-free merging methods and exclude Fisher merging.  
On average, our method outperforms the baseline across all merging techniques, including downstream task performance.  
Notably, the per-task improvement reaches up to $5.92\%$ on TA and $12.88\%$ on TIES.  
Furthermore, compared to previous results, we observe that although larger models generally achieve higher average performance than smaller ones, the performance gain from our method diminishes.  
A possible explanation is that as model size increases, its inherent knowledge also expands, which naturally enhances merging performance, thereby reducing the relative impact of our approach.  

Moreover, we present the results of merging LLaMA3-8B models in~\cref{tab:llama8b-pertask}.
Given the large scale of the model and resource constraints, we evaluate our proposed OSRM using two lightweight merging techniques: TA and TIES.
Our method consistently improves both downstream task performance and the merging effectiveness of TA.
Notably, it outperforms the baseline on seven out of eight datasets when combined with TA, demonstrating its clear advantage.

\paragraph{Discussion.}
First, we observe that the choice of the optimal non-adaptive merging method (i.e., excluding EMR) varies depending on the model.  
For example, TA achieves the best performance on RoBERTa-large, while RegMean outperforms other methods on T5-large.  
Even between two decoder-based models, the most effective merging strategy can differ.  
Second, EMR consistently achieves the highest performance among all methods, performing close to that of individual models.
It implies the superiority of adaptive merging methods.

\subsection{Robustness Analysis}
\label{subsec:robustness}

In this section, we evaluate the robustness of our method to different hyperparameters, including the scaling coefficient $\lambda$, the number of samples $k$, the number of tasks $N$, and the choice of learnable blocks, using RoBERTa-large.

\begin{figure}[t]
    \centering
    \subfigure[TA]{\includegraphics[width=0.45\linewidth]{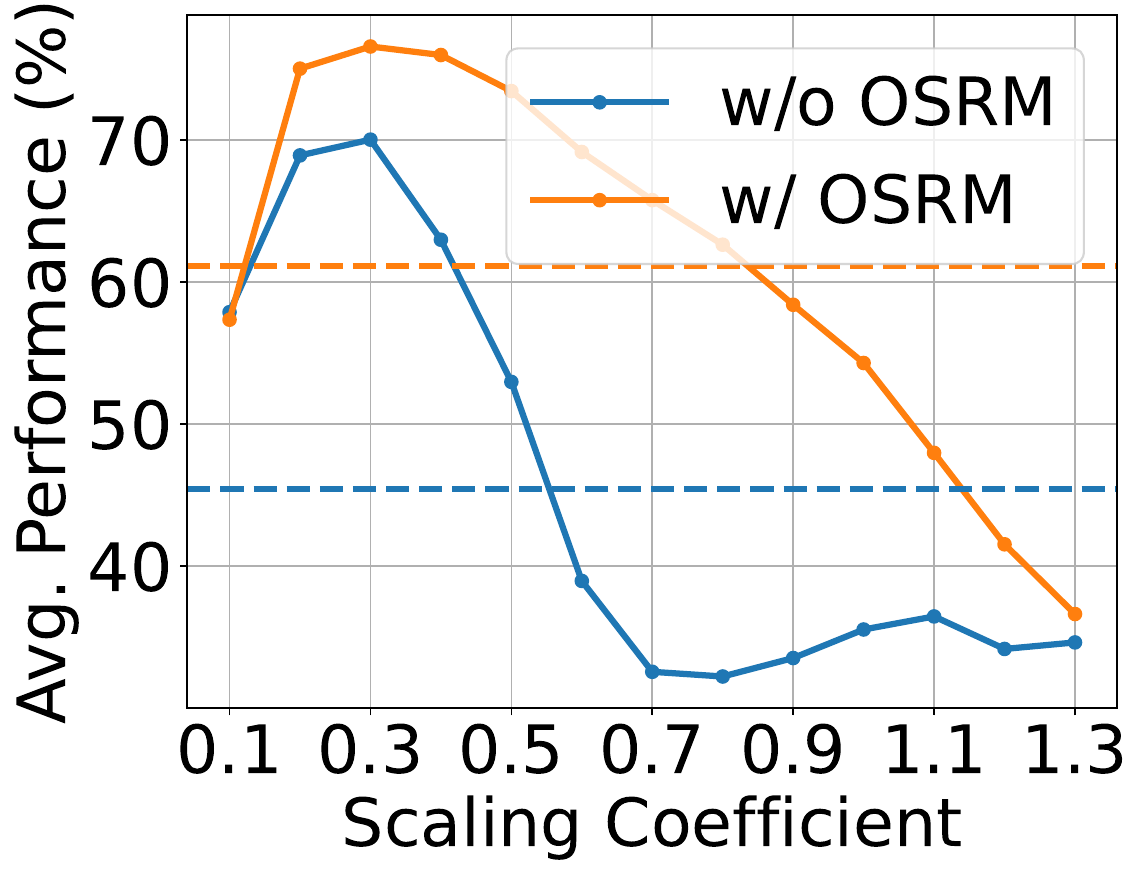}}
    \subfigure[TIES]{\includegraphics[width=0.45\linewidth]{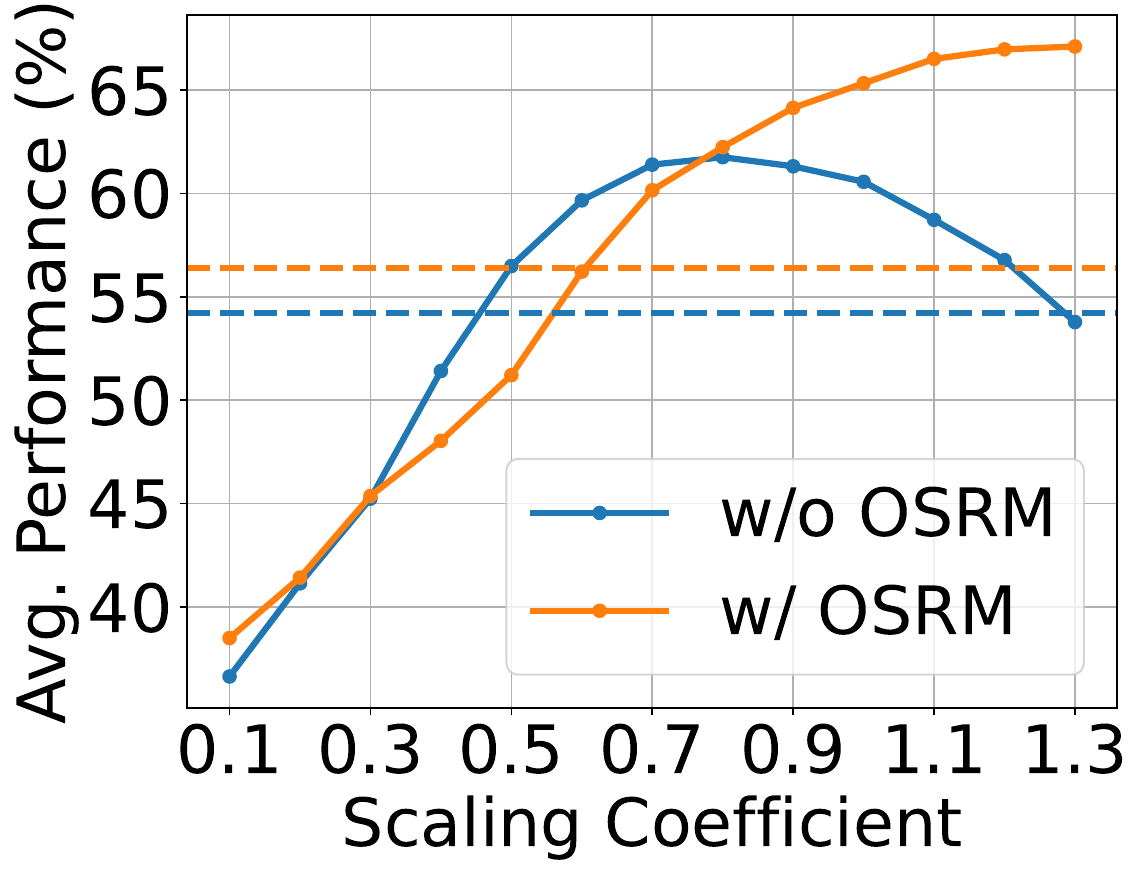}}
    \caption{Effect of scaling coefficients on the performance of TA and TIES merging. Results are averaged across eight datasets. The solid line is the merging performance for each scaling coefficient. The dashed line is the average performance for each method.}
    \label{fig:effect-scaling}
\end{figure}

\begin{table}[t]
    \small
    \centering
    \caption{Effect of the number of samples used to compute \cref{eq:feature concat}. Results are averaged across eight datasets. \textbf{Bold} indicates the best performance for each merging method.}
    \label{tab:effect-num_samples-roberta}
    \resizebox{\columnwidth}{!}{%
    \begin{tabular}{ccccccc}
    \hline
     & 2 & 10 & 100 & 1000 & 5000 & \begin{tabular}[c]{@{}c@{}}w/o OSRM\\ \end{tabular} \\ \hline
    RegMean & 47.66 & \textbf{70.42} & 69.81 & 69.21 & 70.16 & 67.88 \\
    Fisher & 66.19 & 65.22 & \textbf{71.86} & 62.71 & 61.71 & 64.88 \\ 
    EMR & 81.73 & 82.84 & \textbf{83.32} & 81.47 & 78.59 & 81.18 \\ \hline
    \end{tabular}%
    }
\end{table}

\paragraph{Impact of Scaling Coefficient $\lambda$.}
We analyze the effect of different scaling coefficients $\lambda$ on merging performance in~\cref{fig:effect-scaling}.
We focus on TA and TIES, both of which require tuning $\lambda$ before merging.
The values of $\lambda$ are sampled from the range $[0.1, 1.3]$ with a step size of $0.1$.
The results indicate that our method consistently outperforms the baseline and is more robust to variations in $\lambda$.
Specifically, for TA, our approach achieves superior performance across almost the entire range and demonstrates a significantly higher average performance.
Although for $\lambda \in [0.3, 0.7]$, our method is slightly worse than the baseline on TIES, it still yields substantial improvements for other values of $\lambda$ and achieves a higher average overall.
The robustness of our method suggests its practical efficiency, as it does not require fine-grained tuning of the scaling coefficient.

\paragraph{Impact of Sample Size $k$.}
We evaluate the influence of different values of $k$ in~\cref{eq:feature concat} on merging performance.
To minimize the impact of other hyperparameters, such as the scaling coefficient, we focus on RegMean, Fisher, and EMR.
Following~\citep{regmean}, we consider $k \in \{2, 10, 100, 1000, 5000\}$.
Intuitively, increasing $k$ should result in an initialization of $\Tilde{A}$ that is more orthogonal to out-of-task samples, potentially leading to better performance.
However, as shown in~\cref{tab:effect-num_samples-roberta}, the optimal values of $k$ are $10$ and $100$ for RegMean, Fisher, and EMR, respectively.

This phenomenon can be attributed to two key factors.
First, when $k$ is close to $1$, we observe an ill-conditioned latent feature matrix $H$, making the computation of $\Tilde{A}$ more challenging.
Second, as $k$ increases, the knowledge overlap between in-task and out-of-task samples also grows.
In this case, enforcing $\Tilde{A}$ to be orthogonal to these samples may degrade performance by discarding useful shared knowledge.
Due to the complex interplay between fine-tuning procedures, model parameters, and data characteristics, it is challenging to analytically determine the cause of the observed counter-intuitive results. 
We believe the exploration of this situation is interesting.
Despite this, \cref{tab:effect-num_samples-roberta} shows that our method consistently outperforms the baseline with as few as $100$ samples, highlighting its practical applicability.

\begin{figure}[t]
    \centering
    \subfigure[RegMean]{\includegraphics[width=0.45\linewidth]{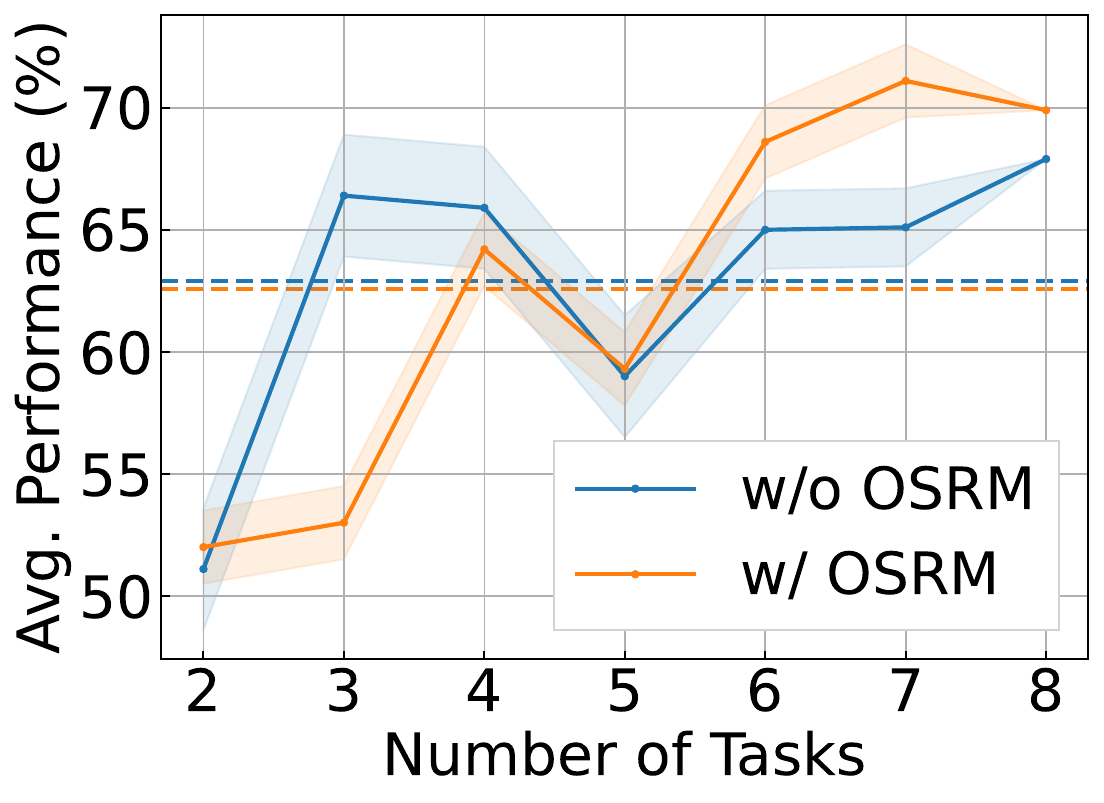}}
    \subfigure[EMR]{\includegraphics[width=0.45\linewidth]{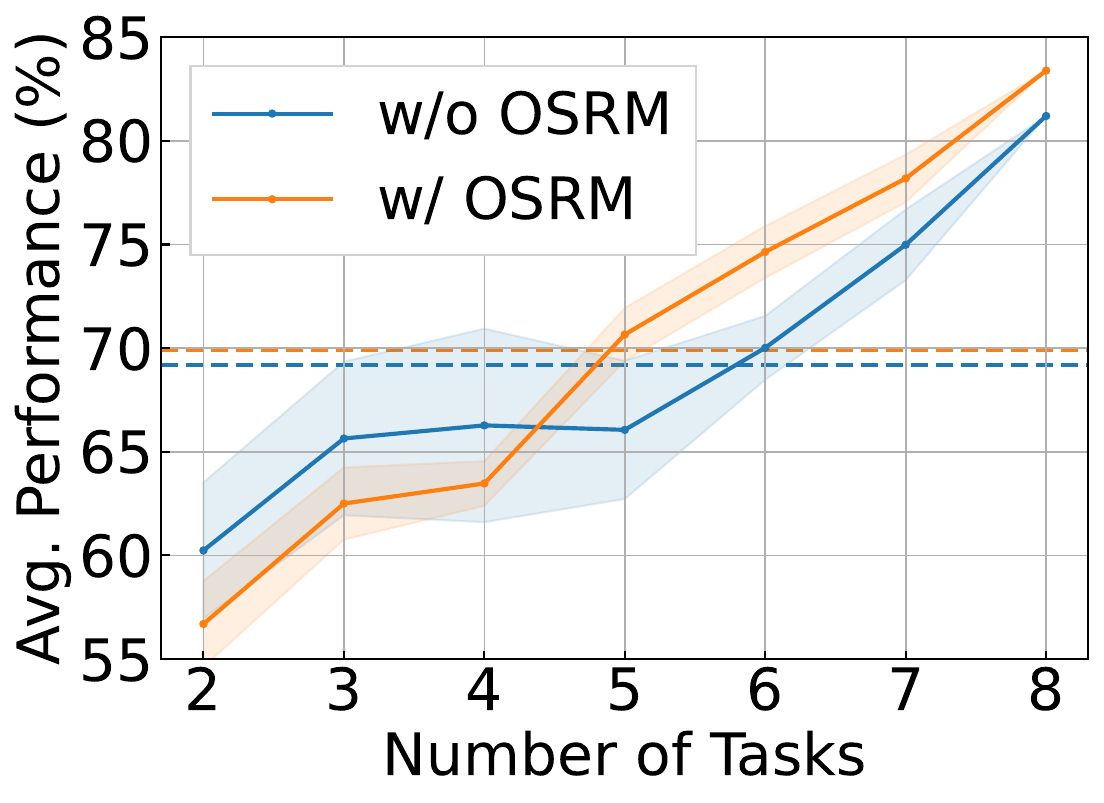}}
    \caption{Performance of merging different numbers of tasks with RegMean and EMR. Results are averaged across all the used tasks and their combinations. The solid line is the merging performance for each number of tasks. The dashed line is the average performance for each method. The shadow is the standard deviation of all the task combinations.}
    \label{fig:effect-numtasks}
    \vspace{-.2in}
\end{figure}

\paragraph{Impact of the Number of Tasks $N$.}
We investigate the effect of the number of tasks $N$ on merging performance in~\cref{fig:effect-numtasks}.
Although the average performance of our method and the baseline remains comparable for small $N$, our approach shows clear advantages when $N$ increases.
Specifically, when $N > 5$, our method begins to outperform the baseline, with improvements becoming more significant as $N$ grows.
This observation suggests that our method enhances merging performance in large-scale settings, demonstrating its scalability.
Moreover, our method has a significantly smaller standard deviation than the baselines, indicating its robustness to task selection.

\begin{table}[t]
\centering
\caption{Effect of different learnable blocks. Results are averaged across eight datasets. \textbf{Bold} indicates better performance.}
\label{tab:effect-ft-blocks}
\resizebox{\columnwidth}{!}{%
\begin{tabular}{cc|cccccc}
\hline
 & OSRM & Individual & TA & RegMean & Fisher & TIES & EMR \\ \hline
\multirow{2}{*}{Q, K, V} & No & \textbf{87.81} & 71.24 & 67.89 & 65.60 & \textbf{51.61} & 78.56 \\
 & Yes & 87.76 & \textbf{75.85} & \textbf{67.95} & \textbf{67.81} & 49.90 & \textbf{82.16} \\ \hline
\multirow{2}{*}{Q} & No & \textbf{86.67} & 71.95 & 43.90 & 58.63 & \textbf{41.63} & \textbf{79.64} \\
 & Yes & 86.26 & \textbf{73.83} & \textbf{52.55} & \textbf{60.04} & 40.78 & 78.80 \\ \hline
\end{tabular}%
}
\end{table}

\paragraph{Impact of Learnable Blocks.}
We evaluate the effectiveness of our method with various learnable blocks in~\cref{tab:effect-ft-blocks}.
While the main experiments in~\cref{subsec:main results} focus on fine-tuning the query (Q) and value (V) blocks, we further investigate the impact of fine-tuning all three blocks—query (Q), key (K), and value (V)—as well as fine-tuning the query blocks alone. Although our method introduces a slight degradation in downstream task performance, it consistently outperforms baseline approaches when integrated with TA, RegMean, and Fisher merging strategies. Additionally, we observe that fine-tuning Q, K, and V yields better performance than fine-tuning Q alone, suggesting that increasing the number of learnable blocks contributes to more effective model merging.

\subsection{Extension to Merging Existing LoRA Modules}
\label{subsec:extension-post-merging}
While there are some cases where the LoRA modules are obtained externally, such as HuggingFace checkpoints, we extend our method to merging existing LoRA modules by decomposing the learned weight with the OSRM-based solution $\Tilde{A}$.
Specifically, given a set of learned weights $\{\Delta W_t\}^N_{t=1}$, we use the analytically-derived $\Tilde{A}_t$ to decompose and recover $\Delta W_t$ by using $\hat{B}_t = \min_B \|B\Tilde{A}_t - \Delta  W_t\|_F$ and $\Delta \hat{W}_t = \hat{B}_t \Tilde{A}_t$.
Then we merge the recovered $\{\Delta \hat{W}_t\}^N_{t=1}$.
While this decomposition may suffer from degradation of the downstream task performance, the recovered $\Delta \hat{W}_t$ can still maintain the property as shown in~\cref{eq:objective}.
See~\cref{app:extension-post-merging} for results.

\section{Conclusion}
We present a novel approach to address performance degradation when merging LoRA-based models. 
By constraining the LoRA subspace before fine-tuning, our method decreases harmful output shifts arising from data and parameter interference. 
Empirical results on eight datasets show that this approach substantially improves upon existing merging strategies across multiple LMs.

\section{Limitations}
While our approach shows significant performance improvement in model merging, some limitations should be discussed.
First, similar to previous works, our method relies on the identical model architecture, which limits its applicability across different model types.
Second, since we only focus on LoRA models, our method cannot be applied to merging fully fine-tuned models.
Future works could further investigate the potential application of our method on models with different architectures or fully fine-tuned.

\section*{Acknowledgments}
We thank anonymous reviewers for their helpful comments.
This material is based in part upon work supported by the National Science Foundation under Grant IIS-2212174, National Institute of Aging (NIA) 1RF1AG072449, National Institute of General Medical Sciences (NIGMS) 1R01GM145700.

\bibliography{main}

\appendix

\section{Proof.}
\label{app:proof}
We prove that~\cref{eq:solution} is an analytical solution to the problem~\cref{eq:objective}.

\begin{proof}
    Let $S=\frac{1}{k-1}H_1^\top H_1$ be the covariance matrix of $H_1$, which is symmetric and positive semi-definite.
    Then 
    \begin{equation*}
        \|AH_1^\top\|_F^2
        = \tr(AH_1^\top H_1 A^\top)
        = (k-1)\tr(ASA^\top).
    \end{equation*}
    The eigendecomposition on $S$ yields
    \begin{equation*}
        S = V \Lambda V^\top,
    \end{equation*}
    where $V \in \R^{n \times n}$ is orthogonal and $\Lambda = \diag(\lambda_1,...,\lambda_n)$ is a diagonal matrix with non-negative entities in a descending order.
    Thus,
    \begin{equation*}
        \tr(ASA^\top) = \tr(AV\Lambda V^\top A^\top).
    \end{equation*}
    Let $M = AV \in \R^{m \times n}$.
    Since $AA^\top=I$ and $VV^\top=I$, it follows that $MM^\top=I$ and 
    \begin{equation*}
        \tr(ASA^\top) = \tr(M\Lambda M^\top).
    \end{equation*}
    Considering $M=AV$, the solution to the problem $\argmin_A \tr(M \Lambda M^\top)$ is spanned by the eigenvectors of $S$ associated with the smallest eigenvalues, i.e., $\Tilde{A}_2 = V^\top_{:, n-r:n}$, where $V_{:, n-r:n}$ is the last $r$ eigenvectors associated with the $r$ smallest eigenvalues.
\end{proof}

\section{Dataset Details}

The GLUE benchmark~\footnote{https://huggingface.co/datasets/nyu-mll/glue} is widely used for general language understanding evaluation.
It consists of eight English datasets.
We show the details of the datasets we use in~\cref{tab:dataset detail}.

\begin{table*}[t]
    \small
    \centering
    \caption{Dataset details in the GLUE benchmark. Acc. and cc. mean accuracy and correlation coefficient, respectively.}
    \label{tab:dataset detail}
    \resizebox{.7\textwidth}{!}{%
    \begin{tabular}{ccccc}
    \hline
    Dataset & \#Train (K) & \#Val (K) & \#Test (K) & Metric \\ \hline
    CoLA & 8.55 & 1.04 & 1.06 & Matthews cc. \\
    MNLI & 393 & 9.82 & 9.8 & Acc. \\
    MRPC & 3.67 & 0.408 & 1.73 & Acc. \\
    QNLI & 105 & 5.46 & 5.46 & Acc. \\
    QQP & 364 & 40.4 & 391 & Acc. \\
    RTE & 2.49 & 0.277 & 3 & Acc. \\
    SST2 & 67.3 & 0.872 & 1.82 & Acc. \\
    STSB & 5.75 & 1.5 & 1.38 & Avg. of Pearson and Spearman cc. \\ \hline
    \end{tabular}%
    }
\end{table*}

\section{Model Details}

We use three language models, including RoBERTa-large~\footnote{https://huggingface.co/FacebookAI/roberta-large}, T5-large~\footnote{https://huggingface.co/google-t5/t5-large}, and Llama3.2-1B~\footnote{https://huggingface.co/meta-llama/Llama-3.2-1B}, and one large language model Llama3.2-3B~\footnote{https://huggingface.co/meta-llama/Llama-3.2-3B} in our experiments.
\cref{tab:model detail} shows the details of used models.

\begin{table*}[t]
    \small
    \centering
    \caption{Details of used models.}
    \label{tab:model detail}
    \resizebox{.4\textwidth}{!}{%
    \begin{tabular}{ccc}
    \hline
    Model & \#Params & Architecture \\ \hline
    RoBERTa-large & 355M & Encoder-only \\
    T5-large & 738M & Encoder-decoder \\
    Llama3.2-1B & 1.24B & Decoder-only \\
    Llama3.2-3B & 3.21B & Decoder-only \\ \hline
    \end{tabular}%
    }
\end{table*}

\section{Hyper-Parameters}
\label{append:hyperparam}

We show the hyper-parameters used to fine-tune language models in~\cref{tab:hyperparam-train}.
Similar to~\citep{roberta}, we use a grid search for the optimal hyper-parameters.
All the experiments are conducted on eight NVIDIA RTX A6000 GPUs.

\begin{table*}[t]
    \small
    \centering
    \caption{Hyper-parameters for fine-tuning language models. 
    }
    \label{tab:hyperparam-train}
    \resizebox{.8\textwidth}{!}{%
    \begin{tabular}{ccccc}
    \hline
    Hyper-param & RoBERTa-large & T5-large & Llama3.2-1B & Llama3.2-3B \\ \hline
    Learning rate & \{3e-5, 2e-4, 4e-4\} & \{3e-5, 2e-4, 4e-4\} & \{1e-4, 2e-4, 4e-4\} & \{2e-4, 4e-4\} \\
    Batch size & \{32, 64\} & \{32, 64\} & \{16, 32\} & \{16, 32\} \\
    Weight decay & \{0, 0.01, 0.1\} & \{0, 0.01, 0.1\} & \{0, 0.1\} & \{0, 0.1\} \\
    Max \#epochs & 10 & 10 & 10 & 5 \\ \hline
    \end{tabular}%
    }
\end{table*}

\section{More Experimental Results}

\subsection{Averaged Results}

We show the averaged performance of each model across all datasets in~\cref{tab:main-roberta-average,tab:main-t5-average,tab:llama1b-avg,tab:main-llama3b-average,tab:main-llama8b-average}, respectively.
Results show that our method outperforms the baseline in almost all settings on average.

\subsection{Results of Extension to Merging Existing LoRA Modules}
\label{app:extension-post-merging}

In~\cref{subsec:extension-post-merging}, we extend OSRM to support the merging of existing LoRA modules, such as externally obtained checkpoints from HuggingFace. 
The corresponding results are reported in~\cref{tab:main-roberta-post-extension-average}. 
When OSRM is applied after fine-tuning, the original LoRA weight matrices cannot be perfectly recovered, resulting in a significant drop in the individual performance of the models.
Nevertheless, the merged performance remains largely preserved, demonstrating the effectiveness of our method.
Preserving individual performance in the post-fine-tuning setting poses an interesting yet non-trivial challenge, which we identify as a promising direction for future work.

\begin{table*}[t]
\small
\centering
\caption{Averaged performance (\%) across eight tasks of RoBERTa-large. "Individual" refers to the metrics of each fine-tuned model on the dataset on which it was trained.
\textbf{Bold} indicates a higher accuracy.
}
\label{tab:main-roberta-average}
\resizebox{0.7\textwidth}{!}{%
\begin{tabular}{c|cccccc|
>{\columncolor[HTML]{EFEFEF}}c }
\hline
OSRM & Individual & TA & RegMean & Fisher & TIES & EMR & Avg. \\ \hline
No & \textbf{88.05} & 70.04 & 67.88 & 64.88 & 61.32 & 81.18 & 72.22 \\
Yes & 87.87 & \textbf{76.59} & \textbf{69.81} & \textbf{71.86} & \textbf{66.65} & \textbf{83.32} & \textbf{76.02} \\ \hline
\end{tabular}%
}
\end{table*}

\begin{table*}[t]
    \small
    \centering
    \caption{Averaged performance ($\%$) across eight tasks of T5-large. "Individual" refers to the metrics of each fine-tuned model on the dataset on which it was trained.
    \textbf{Bold} indicates a higher accuracy.}
    \label{tab:main-t5-average}
    \resizebox{.7\textwidth}{!}{%
    \begin{tabular}{c|cccccc|
    >{\columncolor[HTML]{EFEFEF}}c }
    \hline
    OSRM & Individual & TA & RegMean & Fisher & TIES & EMR & Avg. \\ \hline
    No & 86.38 & 52.96 & 69.92 & 48.23 & 55.49 & 81.02 & 65.67 \\
    Yes & \textbf{87.01} & \textbf{55.79} & \textbf{73.68} & \textbf{56.13} & \textbf{59.74} & \textbf{82.61} & \textbf{69.16} \\ \hline
    \end{tabular}%
    }
\end{table*}

\begin{table*}[t]
    \small
    \centering
    \caption{Averaged performance (\%) across eight tasks of Llama3.2-1B. "Individual" refers to the metrics of each fine-tuned model on the dataset on which it was trained. \textbf{Bold} indicates a higher accuracy.}
    \label{tab:llama1b-avg}
    \resizebox{.7\textwidth}{!}{%
    \begin{tabular}{c|cccccc|
    >{\columncolor[HTML]{EFEFEF}}c }
    \hline
    OSRM & Individual & TA & RegMean & Fisher & TIES & EMR & Avg. \\ \hline
    No & 84.42 & 65.68 & 41.19 & \textbf{61.06} & 77.85 & 78.12 & 68.05 \\
    Yes & \textbf{85.52} & \textbf{68.58} & \textbf{43.57} & 59.64 & \textbf{78.77} & \textbf{78.85} & \textbf{69.16} \\ \hline
    \end{tabular}%
    }
\end{table*}

\begin{table*}[t]
    \small
    \centering
    \caption{Averaged performance ($\%$) across eight tasks of Llama3.2-3B. "Individual" refers to the metrics of each fine-tuned model on the dataset on which it was trained. \textbf{Bold} indicates a higher accuracy.}
    \label{tab:main-llama3b-average}
    \resizebox{.7\textwidth}{!}{%
    \begin{tabular}{c|ccccc|
    >{\columncolor[HTML]{EFEFEF}}c }
    \hline
    OSRM & Individual & TA & RegMean & TIES & EMR & Avg. \\ \hline
    No & 87.54 & 73.22 & 45.53 & 54.73 & 83.43 & 68.89 \\
    Yes & \textbf{88.33} & \textbf{74.36} & \textbf{45.77} & \textbf{57.28} & \textbf{83.47} & \textbf{69.84} \\ \hline
    \end{tabular}%
    }
\end{table*}

\begin{table*}[t]
    \small
    \centering
    \caption{Averaged performance ($\%$) across eight tasks of Llama3-8B. "Individual" refers to the metrics of each fine-tuned model on the dataset on which it was trained. \textbf{Bold} indicates a higher accuracy.}
    \label{tab:main-llama8b-average}
    \resizebox{.5\textwidth}{!}{%
    \begin{tabular}{c|ccc|
    >{\columncolor[HTML]{EFEFEF}}c }
    \hline
    OSRM & Individual & TA & TIES & Avg. \\ \hline
    No & 87.94 & 63.28 & \textbf{53.01} & 68.08 \\
    Yes & \textbf{88.77} & \textbf{72.75} & 52.69 & \textbf{71.4} \\ \hline
    \end{tabular}%
    }
\end{table*}

\begin{table*}[t]
\small
\centering
\caption{Extension to merging existing LoRA modules. "Post" indicates applying OSRM \emph{after} fine-tuning (c.f.~\cref{subsec:extension-post-merging}). Performance (\%) is averaged across eight tasks of RoBERTa-large. "Individual" refers to the metrics of each fine-tuned model on the dataset on which it was trained.
}
\label{tab:main-roberta-post-extension-average}
\resizebox{0.7\textwidth}{!}{%
\begin{tabular}{c|cccccc|
>{\columncolor[HTML]{EFEFEF}}c }
\hline
OSRM & Individual & TA & RegMean & Fisher & TIES & EMR & Avg. \\ \hline
No & 88.05 & 70.04 & 67.88 & 64.88 & 61.32 & 81.18 & 72.22 \\
Post & 35.07 & 35.26 & 35.46 & 35.11 & 35.17 & 35.14 & 35.2 \\ \hline
\end{tabular}%
}
\end{table*}

\section{Discussion}

\subsection{Elaborate Discussion on Limitations}

As we have briefly discussed limitations in the main text, we further give elaborate discussion on limitations in this section.
First, as LoRA has been widely used for fine-tuning LLMs in many areas such as finance, healthcare, and code generation~\citep{smarter-ft}, there is still a performance gap in merging LoRA-fine-tuned models~\citep{knots}. Thus, many methods have been proposed to improve the merging performance of models specifically fine-tuned with LoRA~\citep{knots,partial-linearization,lora-soups}.
Second, the identical model architecture is a common assumption in the area of model merging, such as our baselines, and has many realistic applications, such as one-shot FL~\citep{ta-oneshot-fl} and LLM agents~\citep{cycleqd}. We also believe model merging under various architectures is an interesting problem, but out of this paper’s scope.

\subsection{Comparison with Multi-task Learning}

There are two main differences between OSRM-based training followed by merging and standard multi-task learning (MTL)~\citep{mtl-moo,multitask}.
First, standard MTL is a data-collecting paradigm while merging with OSRM is a model-collecting paradigm. 
Specifically, standard MTL requires the samples from all datasets to be collected together to train a multi-task model. 
Instead, OSRM allows each data source to train its own model and merges the models together.
Second, while standard MTL requires the collection of all the original samples, OSRM only requires a small number of latent features from each task (100 samples per task in our experiments).

Compared to MTL, our method can be applied to cases where full data access has memory issues or privacy concerns, such as model merging and one-shot FL~\citep{ta-oneshot-fl}, or other paradigms such as training LLM agents~\citep{cycleqd}.

\end{document}